# Epically Powerful: An open-source software and mechatronics infrastructure for wearable robotic systems

*Jennifer K. Leestma[1,2,3], *Siddharth R. Nathella[1], *Christoph P. O. Nuesslein[1,2], Snehil Mathur[4,5], Student Member, IEEE, Gregory S. Sawicki[1,2,6], Member, IEEE, and Aaron J. Young[1,2], Senior Member, IEEE

[*]These authors contributed equally to this work
[1]George W. Woodruff School of Mechanical Engineering, Georgia Institute of Technology, Atlanta, GA 30332 USA
[2]Institute for Robotics and Intelligent Machines, Georgia Institute of Technology, Atlanta, GA 30332 USA
[3]School of Engineering and Applied Sciences, Harvard University, Cambridge, MA 02138 USA
[4]School of Electrical and Computer Engineering, Georgia Institute of Technology, Atlanta, GA 30332 USA
[5]School of Mechanical Engineering, Rice University, Houston, TX 77005 USA
[6]School of Biological Sciences, Georgia Institute of Technology, Atlanta, GA 30332 USA

Corresponding author: Jennifer K. Leestma (e-mail: jleestma@seas.harvard.edu).

This work was supported in part by the National Science Foundation Research Traineeship: Accessibility, Rehabilitation, and Movement Science (NSF NRT ARMS) Program Award 1545287, National Science Foundation Graduate Research Fellowship Program (NSF GRFP) Award 1324585, NIH Director's New Innovator Award DP2-HD111709, Georgia Tech Mechanical Engineering Interdisciplinary Research Fellowship, National Defense Science and Engineering Graduate (NDSEG) Fellowship, Georgia Tech President's Undergraduate Research Award, NSF FRR grant 2233164, and NSF TIPS grant 2202862.

**ABSTRACT** Epically Powerful is an open-source robotics infrastructure that streamlines the underlying framework of wearable robotic systems – managing communication protocols, clocking, actuator commands, visualization, sensor data acquisition, data logging, and more – while also providing comprehensive guides for hardware selection, system assembly, and controller implementation. Epically Powerful contains a code base enabling simplified user implementation via Python that seamlessly interfaces with various commercial state-of-the-art quasi-direct drive (QDD) actuators, single-board computers, and common sensors, provides example controllers, and enables real-time visualization. To further support device development, the package also includes a recommended parts list and compatibility guide and detailed documentation on hardware and software implementation. The goal of Epically Powerful is to lower the barrier to developing and deploying custom wearable robotic systems without a pre-specified form factor, enabling researchers to go from raw hardware to modular, robust devices quickly and effectively. Though originally designed with wearable robotics in mind, Epically Powerful is broadly applicable to other robotic domains that utilize QDD actuators, single-board computers, and sensors for closed-loop control.

## I. INTRODUCTION

Wearable robotic controls researchers are often faced with a choice between buying a pre-existing off-the-shelf robotic system or developing their own custom robotic system from the ground up. Off-the-shelf systems have the benefit of well-designed, robust, and ready-to-use physical components along with documentation and technical support [1], [2], [3], [4], [5], [6]. However, these systems often lack customizability and may operate using guarded or proprietary platforms that inhibit full access to mid- and low-level control structures and parameters. On the other hand, developing custom hardware and software enables researchers to tailor systems to their specific application with full access to system controls [7], [8], [9], [10], [11], [12], [13], [14]. However, these endeavors often suffer from time-intensive and expensive development processes and the lack of broader community support and collaboration throughout development and deployment. This tug-of-war between pre-set reliability and flexibility impedes researchers from being able to progress fundamental controls research that is crucial to the advancement of wearable robotic systems.

In recent years, the development of custom wearable robotic systems has been propelled by advances in commercially available sub-components. Specifically, the popularization of quasi-direct drive (QDD) actuators has hugely benefitted the robotics community; these new commercially available QDD actuators are fast, backdrivable, open loop torque controllable, and low-profile “pancake” shaped actuators that are well suited for exoskeletons, prostheses, quadrupeds, and related types of robotic systems [15], [16], [17], [18], [19], [20]. Though these developments have sped up the mechanical design process by enabling and enhancing the quality of custom builds, researchers are often still tasked with developing mechatronic architectures and software interfaces from scratch as well as ensuring cross-component compatibility and integration. These development efforts are time consuming and typically lead to non-standardized setups that contribute to existing research silos. This inhibits groups from easily collaborating, sharing control frameworks, comparing controllers across hardware, or even maintaining control system uniformity within individual lab groups. There is a need for an open-source mechatronics and software infrastructure that is robust, reliable, and modular to unify the controls implementation with these commercial actuators and broader wearable robotic systems.

Recent research efforts have highlighted open-source hardware and software research tools as promising approaches to reduce development time and propel scientific progress. During the rise in popularity of modern-day wearable robotics research, the Flexible, Scalable Electronics Architecture (FlexSEA) offered streamlined and efficient embedded hardware with a software stack for core functions such as motor control and sensor integration [21]. Though this contribution is well-suited for wearable robotic applications, the C-based programming presents a higher barrier-to-entry and does not leverage recent advancements in single-board computers that could enable quicker and simpler prototyping and implementation. With similar motivations, the Open-Source Leg (OSL) was developed as a combined hardware-software platform for developing lower limb prostheses, aimed at standardizing hardware and enabling cross-group comparisons, replication, and collaboration [22]. However, this development is prosthesis-specific, inhibiting use for broader and customizable robotic systems. Similar to the OSL, the OpenExo project streamlines the hardware and software development of upper and lower limb exoskeletons [23]. However, the software architecture is designed to work with specific hardware configurations, and the Arduino-based programming interface deters advanced and data-driven code implementation without an additional computer. Finally, the M-BLUE exoskeleton provides an affordable and modular open-source hardware setup for lower limb exoskeletons [8]. However, the mechanical-only open-source design requires mechatronic and software development. Together, these efforts illustrate the success of open-source efforts lowering the barrier for robotics research, enabling researchers to address fundamental questions more quickly and reliably. While these open-source efforts have advanced the field by offering accessible and specialized solutions, they often lock developers into pre-determined hardware setups. Our goal is to provide an extensible, modular hardware-software infrastructure that supports a wide range of hardware configurations without sacrificing rigor or usability.

In this paper, we present Epically Powerful (named after the Georgia Tech EPIC and PoWeR Labs – the two research groups that collaborated to create the platform), an open-source software and mechatronics infrastructure to accelerate the development of custom wearable robotic systems. The cornerstone of the Epically Powerful package is the software infrastructure to communicate with commercially available actuators via controller area network (CAN) communication, notably CubeMars AK Series, RobStride, and CyberGear actuators. In addition, Epically Powerful contains software packages for analog sensing from various commercial inertial measurement units (IMUs), clocking, visualization, safety monitoring, and other essential plumbing for robotic system operation. Our system leverages single-board computing and a Python-based interface, offering a simple and accessible platform for hardware integration and software development. The installation of Epically Powerful is simple, requiring only a few lines of code to install the package and all dependencies, set up a hotspot, and other tasks needed for immediate operation. The software package is accompanied by extensive documentation, recommended parts list and compatibility guide, and detailed instructions on both software and hardware implementation. This package is designed to be modular and extensible, allowing users to easily swap components, update controllers, and add functionality without needing to reconfigure the essential hardware and software that enables baseline functionality. Epically Powerful enables both novice and expert roboticists to accelerate custom wearable robotic system development without sacrificing rigor, efficiency, or reliability. Full documentation is available at https://gatech-epic-power.github.io/epically-powerful/ and code repository is available at https://github.com/gatech-epic-power/epically-powerful.

## II. CONSTRUCTING AN EPICALLY POWERFUL ROBOT

Epically Powerful is intended to serve as an adaptable mechatronics framework and software stack that enables the rapid development and operation of wearable robotic devices. The hardware and software infrastructure supports modular combination of actuators, single-board computers, sensors, and power sources into one cohesive mechatronic system, facilitating the realization of custom application-specific devices (Fig. 1). We intentionally decouple the core mechatronics and control functionality from domain-specific device configurations, enabling the user to design a physical robot that is best suited for their application. For example, if using Epically Powerful to construct and control a hip exoskeleton, the user can freely develop the human interfacing components that couple the actuators to the user as well as a backpack or other holder for the mechatronics components. The compatibility of all included subcomponents (i.e. actuators, computers, sensors, and power sources) ensures that users can configure a device tailored to their application, rather than being locked into a pre-set hardware configuration.

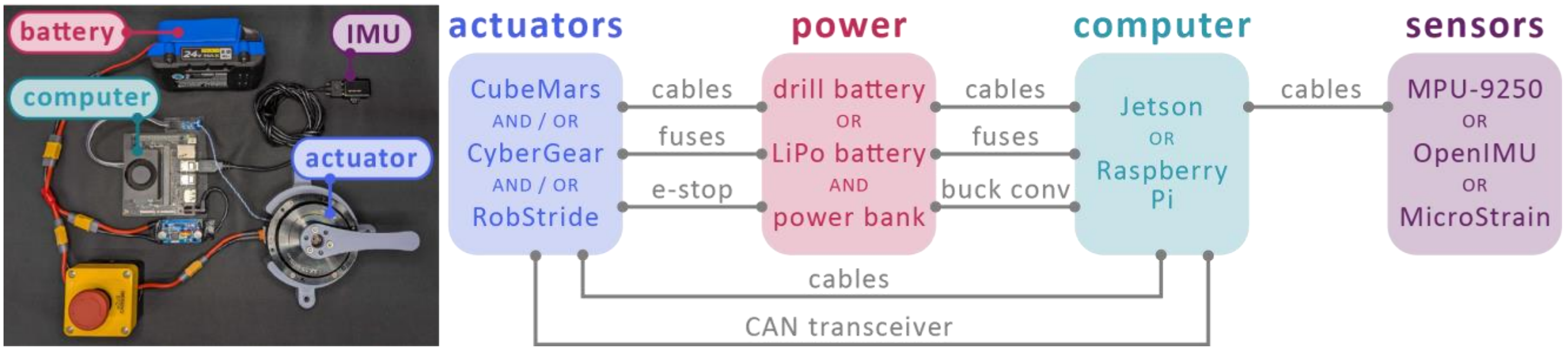

**FIGURE 1. The compatible hardware setup for Epically Powerful is meant to be modular and customizable. The three main components that users need to select are actuators, power source(s), and a computer, with optional sensors. The system is compatible with all CubeMars AK-Series actuators, all RobStride actuators, and the CyberGear Micromotor. A Li-Ion drill battery or LiPo battery can be used to power the actuators and computer, with the computer being optionally and separately powered by a power bank. Users can also choose between an NVIDIA Jetson Orin Nano or Raspberry Pi. Users can interact with common inertial measurement unit (IMU) sensors, alongside native actuator encoders. We include recommendations and setup instructions for all parts needed to interface between the actuators, power, computer, and sensors.**

Epically Powerful includes extensive online documentation and open-source code repository [24], [25]. The documentation discusses part selection, subcomponent setup, mechatronic assembly, actuator configuration, as well as software deployment and testing. Users can select from a suite of commercial QDD actuators, single-board computers (Raspberry Pi and NVIDIA Jetson Orin), sensors, and power source options. The documentation provides further guidance on assembling power and communication subsystems, implementing safety options, configuring devices, and other steps needed to get a robot up and running, supporting reproducibility and rapid development.

The software package, available through the Python package index (PyPI), provides robust functionality for any combination of chosen hardware subcomponents. Crucially, the software package abstracts low-level control operations, enabling the user to focus on mid- and high-level controller development using Python (Fig. 2). The included supporting functions for real-time control operation alongside example controllers lower the development time barrier for both novice and experienced roboticists.

Broadly, Epically Powerful aims to provide users with domain-specific design flexibility while accelerating development with modular, vetted hardware and a robust and adaptable software stack. Although originally designed by our group for the development of wearable robots and related devices, the common use of QDD actuators, single-board computers, and IMU sensors across many areas of robotics allows this package to be useful to the broader robotics community.

## III. HARDWARE COMPONENTS

We designed Epically Powerful's control system to be compatible with a suite of commercially-available, off-the-shelf components, allowing researchers to choose the components that are best suited for their application. We accommodate several QDD actuators that span a large range of torque capabilities applicable for wearable robotics, two classes of single-board computers, various IMU sensors, and power source options (Fig. 1). All computers are compatible with all actuators and sensors and can also be powered using any of the included battery types, enabling users to select components specific to their application. Our ordering guide not only outlines these component options but also highlights key implementation needs across these categories (e.g. battery choices best suited to selected components, needed connectors, etc.). Here, we discuss the hardware components that we have ensured are compatible with Epically Powerful software and that are discussed in our ordering and compatibility guide (Fig. 1).

### *A. ACTUATORS*

The supported and compatible hardware includes QDD actuators, available through CubeMars, RobStride, and CyberGear. At the time of this publication, the compatible actuators provide rated torque capabilities ranging from 1.3-40 Nm, peak torque capabilities ranging from 4.1-120 Nm, with costs ranging from \$136-\$799. These compatible actuators include all CubeMars AK Series actuators, all RobStride Dynamics RobStride Series actuators, and the Xiaomi CyberGear Micromotor. Because the CubeMars AK80-64 is not a QDD actuator and the RobStride 00 does not have a thin, pancake-like profile like the others discussed throughout this work, they are excluded from our parts list though they are also compatible with the Epically Powerful system. We also intend to continually update the package following publication to ensure compatibility with any new state-of-the-art QDD actuator releases.

Each of these actuators contains an encoder, providing position and velocity measurements during actuator operation. Because this is a sensor that is natively included with actuation, encoders are not discussed in the Sensors section of this paper. Similarly, in the code base, encoders are referenced through the *Actuation* class rather than the *Sensing* class.

Our documentation and ordering guide provide lists of various components needed to ensure actuator function and safety. Each set of actuators should be accompanied by a CAN transceiver board that enables CAN communication with the chosen computer. Additionally, actuators can be powered using one of our recommended lithium polymer (LiPo) battery or lithium ion (Li-Ion) drill battery options, along with fuses that protect the system if current is overdrawn. Finally, actuators should also include an emergency stop to ensure user safety. All of these components are included in the recommended parts list and assembly instructions.

### B. SINGLE-BOARD COMPUTERS

Epically Powerful is compatible with both the NVIDIA Jetson Orin Nano series and the Raspberry Pi series of single-board computers running a Linux-based operating system. These are highly capable devices that have been broadly used to control advanced wearable robotic systems [10], [12], [26], [27]. The Jetson Orin Nano provides a higher-end option that is particularly suited for real-time machine learning and vision processing due to its onboard NVIDIA graphics processing unit (GPU) with Compute Unified Device Architecture (CUDA) cores. The Raspberry Pi has reduced real-time machine learning capacity but is an excellent lower-cost, lower-power, and lower profile alternative for diverse robotic applications. We intend to ensure Epically Powerful compatibility with future hardware and software updates to these single-board computer lines.

In the ordering and compatibility guide, we provide various power options for these computers. Computer power can either be shared with actuators or be powered separately. If power is shared, we provide a recommended buck converter that enables the battery's voltage to be stepped down to a level that is appropriate for the computer along with a fuse that protects the system from overdrawn current. If power is separate, we provide a recommended power bank.

### C. POWER SOURCES

Li-Ion drill or LiPo batteries are used to power the actuators in the system, as well as the computer if the user chooses a shared power source. In the ordering and compatibility guide, we recommend suitable batteries for each of the actuators that are compatible with Epically Powerful, including various Li-Ion drill battery and LiPo battery options. Li-Ion drill batteries are comparably low-maintenance, safe, easily accessible, and durable, and thus we strongly recommend them over LiPo alternatives, though they suffer slightly from being bulkier and heavier than their LiPo counterparts. LiPo batteries are comparably more power dense, allowing them to be lighter in weight. However, LiPo batteries are also susceptible to catching fire, so we only recommend them if the user group is familiar with and equipped with proper LiPo charging equipment and storage. Computers can be powered by sharing and stepping down the actuator power or by using a separate power bank. We recommend components for both options and discuss setup instructions in the documentation (https://gatech-epic-power.github.io/epically-powerful/).

### D. SENSORS

To complement the on-device computation and actuation, which includes encoder sensors, Epically Powerful currently supports three IMU types: the MicroStrain series (HBK MicroStrain, Williston, VT, USA), OpenIMU series (ACEINNA, Tewksbury, MA, USA) and the MPU-9250 unit (TDK InvenSense, San Jose, CA, USA). The MicroStrains provide onboard functionality to filter and derive orientation from raw sensor data, making them suitable for higher-end sensing tasks where orientation is valued. The OpenIMUs offer similar features including orientation algorithms that, like the MicroStrains', are commercially supported and easily accessible. The MPU-9250s offer more basic functionality, though their affordability makes them desirable for more budget-conscious implementations. It is important to note that all three options provide essentially the same basic accelerometer and gyroscope readings, with little difference in quality between sensors. As newer models in these sensor product lines become available, we intend to integrate them, along with additional sensor types, into Epically Powerful.

The ordering guide includes not only each sensor, but also any needed or beneficial peripherals. The MicroStrains communicate via universal asynchronous receiver-transmitter (UART) cables provided with each unit, many of which can be connected over a single USB distributor. Similarly, the OpenIMUs, which use the CAN protocol like our actuator options, can be connected over the same CAN bus. By contrast, the MPU-9250 IMUs, which use the inter-integrated circuit ($I^2C$) protocol for communication, can be configured with multiplexing boards to enable more than two sensors to be connected on the same $I^2C$ bus. Notably, none of the IMU communication methods conflict with the actuators' CAN communication, ensuring that all sensors are compatible with any actuation setup. Any configuration of these sensors can be set up together on Epically Powerful-supported computers, making multi-sensor configurations straightforward to implement.

## IV. SOFTWARE PACKAGE

The Epically Powerful software package enables a seamless, Python-based user application programming interface (API) for bidirectionally communicating with actuators, querying data from sensors, visualizing data in real-time, clocking, and recording data (Fig. 2). Specifically, the actuation portion of the software package abstracts away complex low-level actuator communication handling and operation monitoring, enabling researchers to focus their effort on mid- and high-level controller development. Broadly, the package is designed to handle the essential plumbing of a robotic control system, while enabling users to readily interact with key operations that may vary across use cases. It should be noted that, if any specific application does require the alteration of the code's background operation, the code is fully accessible and editable for customization.

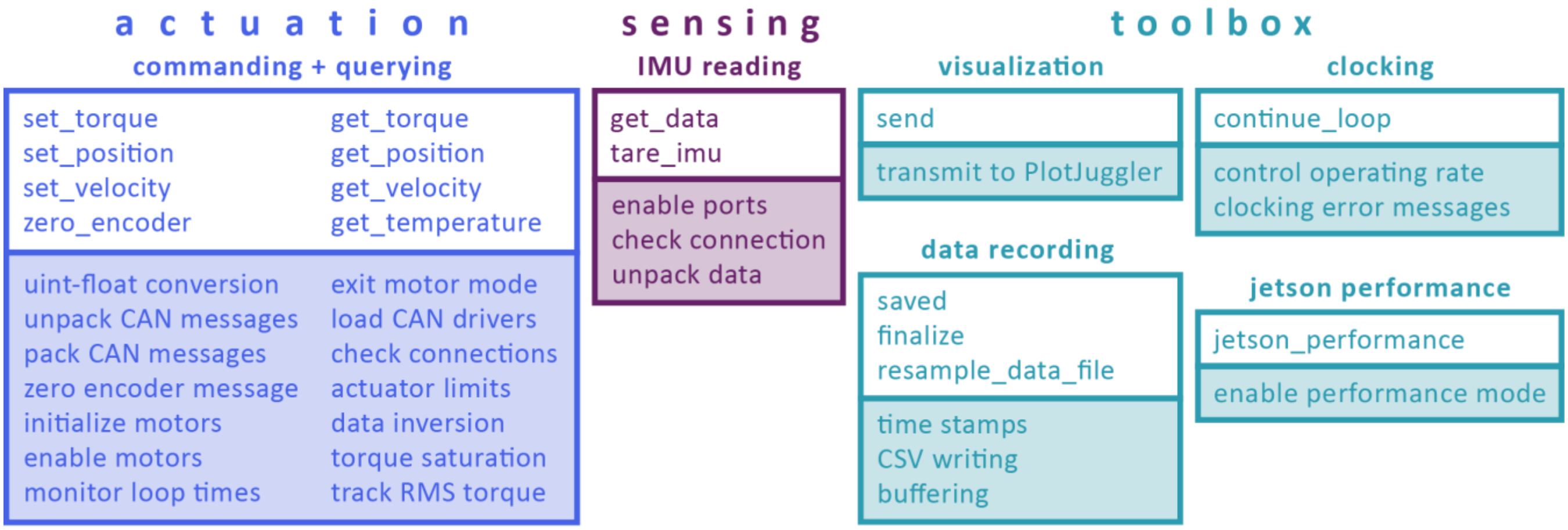

**FIGURE 2.** **An overview of the Epically Powerful software architecture and functionality. Users are intended to regularly interact with frontend functionality (white boxes), which has a consistent stylistic implementation across device types and code function. The backend functionality (shaded boxes) provides underlying granular structure needed to support simple and robust frontend use.**

### A. ACTUATION

Epically Powerful provides a flexible and robust framework for interfacing with a variety of actuators via CAN communication. The main functionality is handled using *ActuatorGroup* objects, which are used to initialize and manage a set of actuators. These can include a mix of CubeMars, RobStride, and CyberGear actuators. To initialize an *ActuatorGroup*, the user constructs the object by specifying the type of each actuator (e.g. "*AK80-9*") and its corresponding CAN ID. Actuator-specific limits (position, velocity, torque, proportional-integral-derivative (PID) gains) are included for each compatible actuator and are used to ensure proper communication behind the scenes. Supporting modules add functions such as checking connection status, automatically zeroing torque when the system is disabled, warning the user when rated torques are being exceeded, optionally saturating torque to below rated limits, and more (Fig. 2).

The actuation class manages initialized actuators, verifies connectivity, sends commands to actuators, and queries the actuators' current states. Actuators can be driven using desired torque, position, or velocity, with an optional current control function for CubeMars actuators. Actuators can also be queried to provide torque, position, velocity, and temperature state. We implemented safety and monitoring functionality, such as tracking root mean squared torques over a 20 second window with an optional to automatically limit torque to prevent overheating of the actuators. Supporting functions handle low-level bitwise conversion and CAN message packing and unpacking.

The actuator control framework is implemented on top of the *python-can* library, and by default is designed to use the SocketCAN protocol of the Linux specification [28], [29]. Working directly with different CAN bus interfaces is often cumbersome, requiring handling of low-level differences in message formats and parsing. Epically Powerful abstracts this process by handling lower-level interfacing with the *python-can* library, enabling users to interact with systems in a consistent way across different CAN bus interfaces. Our approach allows users to pass arguments to the *ActuatorGroup* objects, allowing the same message handling and sending functions to be used regardless of the underlying interface and actuator type.

Users should note that CubeMars actuators have two operating modes, referred to as "MIT" and "Servo" mode, that can be initialized and used via Epically Powerful (Fig. 3). Our approach uses MIT mode by default, as that is our recommendation for most users. MIT mode accepts any combination of position, velocity, and/or torque commands and attempts to drive the actuators directly without utilizing cascaded loops (Fig. 3A). In comparison, Servo mode utilizes separate loops for position, velocity, and current (Fig. 3B,C,D), driven by gains that can only be set via the CubeMars R-Link software. Servo mode does, however, allow for direct control of current, which yields slightly higher maximum torque outputs than the MIT mode and may be desirable for specific applications. Using direct current control is also possible for the RobStride and CyberGear actuators, but is not currently included in our implementation.

### B. SENSING

The Epically Powerful package supports three types of IMUs. Epically Powerful wraps around HBK MicroStrain Communication Library (MSCL) package to communicate with the MicroStrain IMUs and directly initializes MPU-9250 and OpenIMU units [30]. Each IMU type is managed by its own class with low-level implementation handled in the background by Epically Powerful, allowing users to employ consistent commands to query linear acceleration, angular velocity, magnetometer, and temperature data from each IMU type. Additionally, users can query the MicroStrain's direct orientation estimates and utilize their various forms of configurable signal filtering. Packages and setup steps to configure the computer to interface with these sensors are either automatically installed on package setup or included in the Epically Powerful documentation.

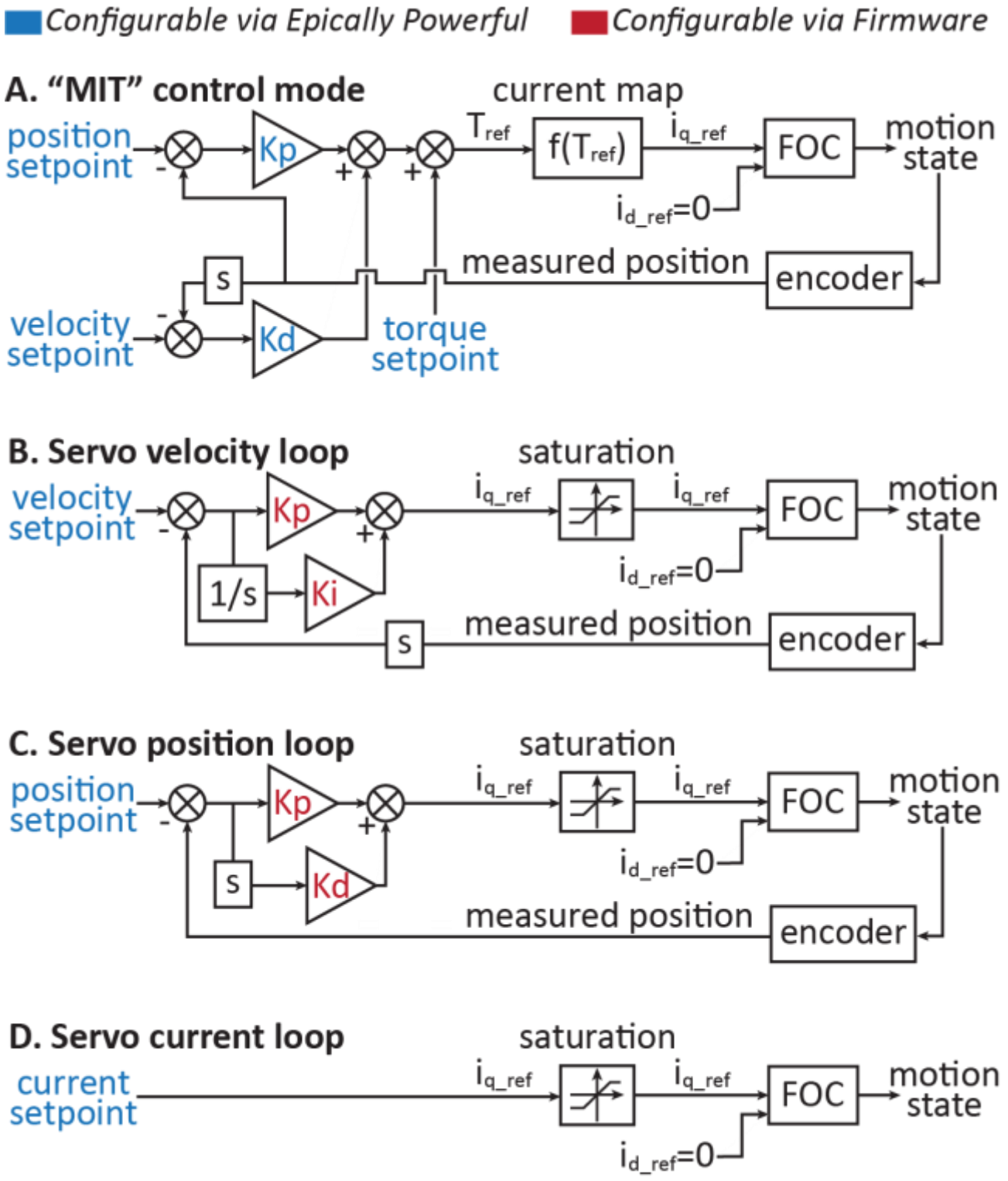

**FIGURE 3.** **Control loops describing the four available modes. A) "MIT" control mode allows for compound control of torque, position, and velocity commands, with Kp and Kd parameters. Epically Powerful implements this for the CubeMars, CyberGear, and RobStride actuators B) Velocity loop controls around a reference velocity setpoint. Kp and Ki are configurable through firmware setup. C) Position loop controls around a reference position setpoint. Kp and Kd are configurable through firmware setup. D) Current loop controls around a reference current setpoint. All methods utilize field-oriented control (FOC). B-D are implemented in Epically Powerful for CubeMars only.**

## C. TOOLBOX

The Epically Powerful toolbox provides various functions to support the main actuation and sensing functionality during real-time control. These additions include code for regulating operating frequency, visualizing data in real time, and recording data, alongside example scripts that show sample implementations of various Epically Powerful functions.

### 1) CLOCKING

Epically Powerful includes a clocking function to ensure fixed-rate operation of control loops. The clocking functionality is implemented using Cython, allowing for high precision sleeping via direct calls to C functions in the operating system. The clocking uses a "scheduling" approach, which cleverly lengthens and shortens each loop so that over time the average loop time is very close to the target loop time, achieving more consistent clocking performance than other approaches. It should be noted that maximum operating frequencies will be determined by the contents of the control loop. We found that operating frequencies are largely limited by the number of actuators included and the chosen computer.

We performed tests to determine the maximum operating rate based on number of actuators and computer. This is the operating rate at which the CAN buffer can accept data, which is separate from the execution time required for contents in a typical control loop. The maximum operating rate is meant to act as an upper bound, but control operation will likely be limited by computations or other functions in the main loop. We used both a Jetson Orin Nano and Raspberry Pi 3B as well as various CubeMars AK Series actuators, testing all combinations between one and eight actuators. We initialized testing with a search space between 50 to 9000 Hz and performed binary search until the search space resolution fell below 100 Hz. Actuators were run for 60 seconds at each frequency, with the tested frequency being eliminated if the computer errored and failed to retain the operating rate. The final chosen frequency was validated for five minutes to confirm longer duration operation. Figure 4 shows the results from both computers across the various actuator counts, along with reference lines representing the approximate operating frequency of various state-of-the-art research devices [8], [9], [10], [22], [23], [31], [32]. If additional time-intensive operations are added to a control loop (e.g. machine learning models), we encourage characterizing the added operation times and modifying these recommended values.

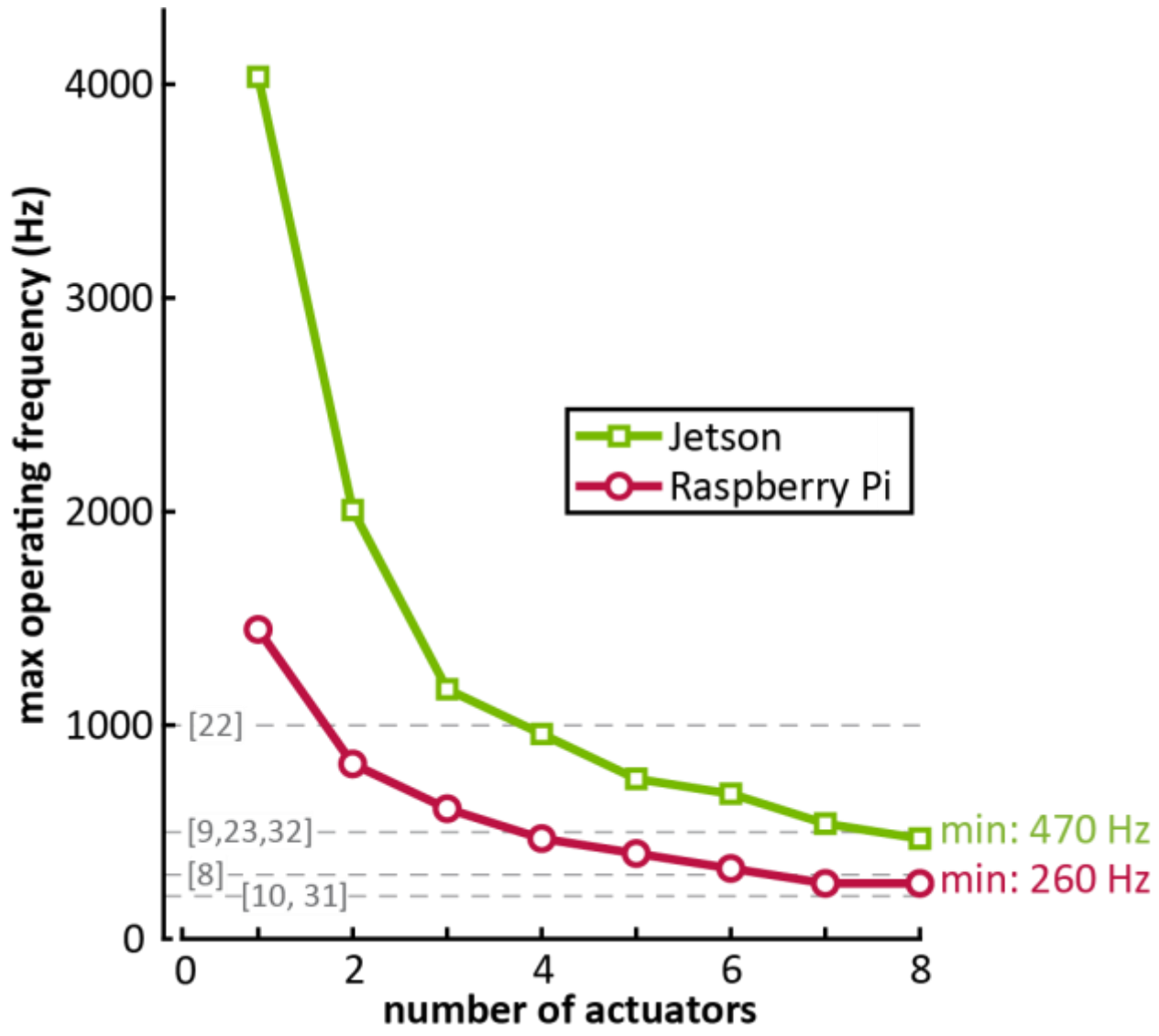

**FIGURE 4.** **The maximum operating frequencies on each computer for various actuator counts. We tested up to eight actuators, with the minimum operating frequency of the NVIDIA Jetson Orin Nano Super being 470 Hz and the Raspberry Pi 3B+ being 260 Hz. Dashed horizontal lines show the approximate operating frequencies of several reference devices by citation. Lines are at 1000Hz, 500Hz, 300Hz, and 200Hz.**

### 2) DATA VISUALIZATION

Epically Powerful is designed to be compatible with PlotJuggler, a widely used visualization tool that is commonly deployed alongside middleware, such as the Robot Operating System (ROS) [33], [34]. To simplify the implementation process, Epically Powerful provides a simplified interface for sending data to PlotJuggler via a user datagram protocol (UDP) socket with messages encoded as JavaScript Object Notation (JSON) strings. This allows the user to send a single line per loop, updating PlotJuggler with the most recent controller data. Importantly, PlotJuggler can be used to view data on a different device (e.g. an external laptop or tablet to view data during experiments) than the one running an Epically Powerful controller, with installers available for Windows and Linux, and source builds for Mac OS.

### 3) DATA RECORDING

We also include a data recorder that seamlessly writes to text files, which are comma delimited by default (i.e. CSV files). The file is written to throughout recording, limiting data loss if errors occur during recorded trials. By default, the data recording buffers 200 frames of data, dumping this data to the written file through a background thread once that buffer fills (e.g. data will be added to the written file once per second for a 200 Hz controller).

## V. DISCUSSION

Epically Powerful provides a robust, modular, customizable, and accessible framework to build and control wearable robotic systems. In recent years, advances in commercially available system components have caused the wearable robotics field to slowly and independently unify how robots are constructed for controls-focused research. Despite this convergence towards similar hardware, the incorporation of subsystems and underlying software architecture are still largely developed in time-intensive silos. We created Epically Powerful to enable the rapid development and deployment of the common core functions of any wearable robotic system that can sense, think and act in real time. Drawing on an inventory of the most commonly used components in wearable robotics systems today, we provide an open-source suite of recommendations for hardware component selection and assembly along with a modular software architecture that can seamlessly interact with each subsystem. Thus, Epically Powerful is meant to enable researchers to go from scientific ideation to an operational robotics platform quickly, with fewer redesigns and debugging sessions along the way.

In comparison to previously published open-source efforts, Epically Powerful employs a modular architecture that enhances both robustness and versatility. For example, Epically Powerful can function as the integration and software backbone for open-source hardware designs, such as the Michigan M-BLUE exoskeleton [8], or act as the foundation for user-customized physical architectures (Fig. 5). It is specifically designed so that users can make hardware innovations, rather than being fixed to a specific hardware setup. Users can select from a slew of actuators, computers, power sources, and sensors, with all combinations being compatible with each other and the underlying software architecture. Researchers benefit from this modularity by being able to leverage vetted key hardware and software without being locked into a broader physical robotic architecture. Though this package was originally designed with wearable robots in mind, the common use of QDD actuators and single-board computers in various areas of robotics ensures that this system can be broadly adopted (Fig. 5).

Two other open-source frameworks, the OSL and OpenExo, offer similar feature sets but have different design goals and scopes than Epically Powerful [22], [23]. The OSL package is most similar to Epically Powerful, as it features a relatively simple Python interface and supports various sensing and actuation systems. However, the OSL package is primarily intended for use with the OSL prosthetic hardware. Thus, tutorials and setup instructions are largely centered around the hardware intended for that prosthetic device. Our framework is instead designed to support arbitrary device setups and accordingly supports a broad range of sensing and actuation options, with complete setup instructions for all supported systems. OpenExo, on the other hand, is built on a fundamentally different approach, with an integrated hardware and software design suite. Thus, OpenExo is best suited for systems that rely primarily on its natively supported hardware and software components. Epically Powerful may be particularly useful for researchers seeking to develop custom hardware configurations that extend beyond OpenExo's native hardware support.

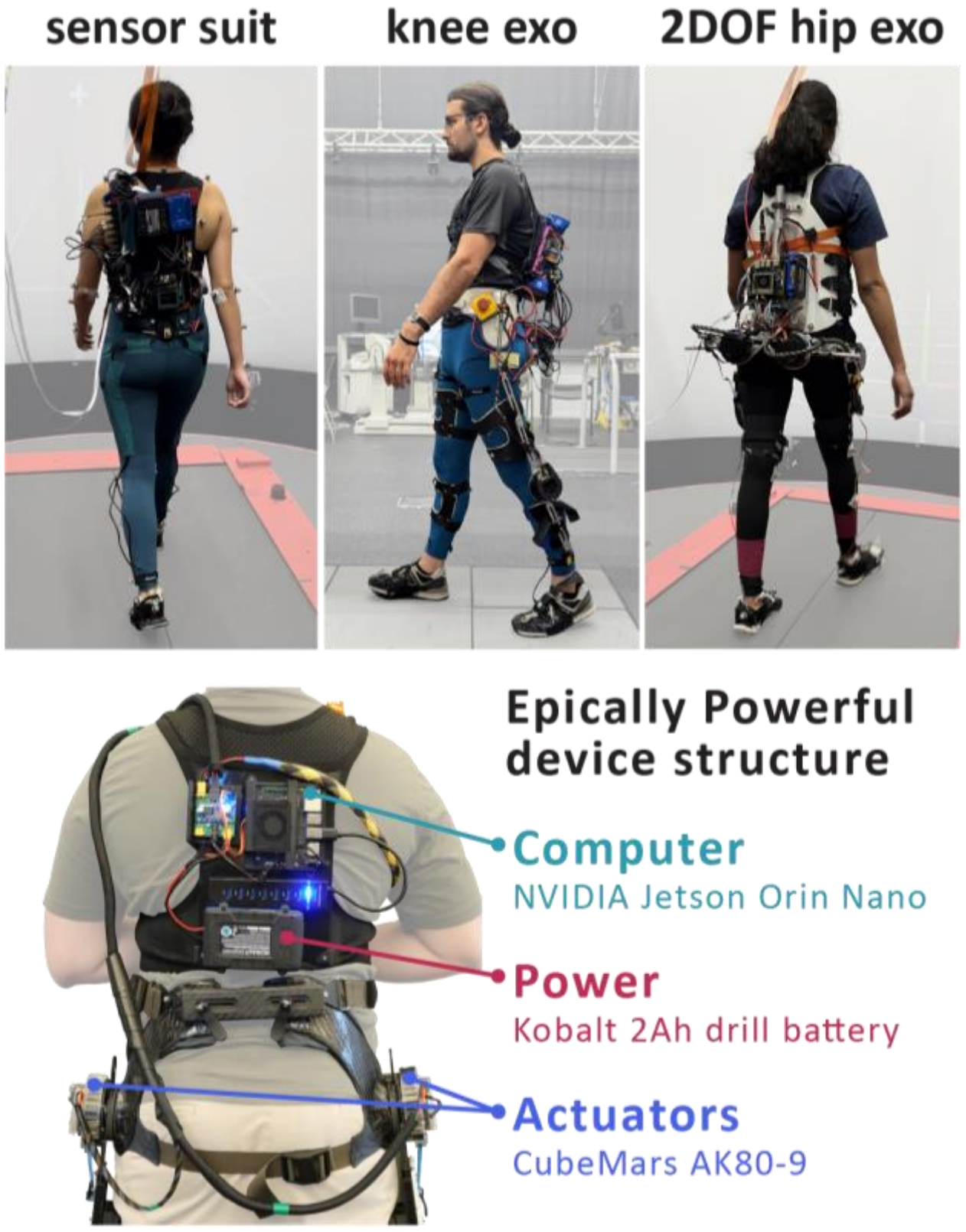

**FIGURE 5.** Epically Powerful serves as the core mechatronics infrastructure supporting a range of customizable mechanical architectures. We have used Epically Powerful to develop and operate both sensor-only suits and a variety of exoskeleton devices. The core subcomponents of the system – actuators, single-board computers, and power sources – are standard across many areas of robotics, ensuring that Epically Powerful provides utility for both wearable and general robotics applications.

In addition to the modular hardware system, Epically Powerful presents an accessible and adaptable software architecture. The Python-based programming interface lowers the barrier for use while also easily integrating with machine learning architectures for more advanced applications. Specifically, the majority of wearable robotic controls research focuses on developing mid- and high-level controllers rather than low-level controllers. This package handles low-level control, which can otherwise be a frustrating and time-consuming aspect of system development, particularly for novice roboticists. Along these lines, the Epically Powerful code wrapper enables users to interact with all components in a consistent style regardless of underlying differences in low-level communication protocols and information structures. The open-source nature of the software package also gives users full control to alter or append to the code base for application-specific contexts, such as incorporating custom sensing peripherals. The software architecture is generalizable, enabling users to easily add, swap, or update device components.

The maintenance of this package is an ongoing effort that the EPIC and PoWeR Labs at Georgia Tech have and intend to continue, including addressing any system bugs, evaluating compatibility with new components, and making updates to accommodate recent product releases. Our goals for future development of this package include integrating non-native driver boards, enabling the implementation of custom motor thermal models, and extending implementation to include a broader array of commonly used sensors. At the time of this publication, Epically Powerful is the backbone of eleven different wearable robotic systems across four universities (Appendix A). Our goal is that researchers in both wearable and broader areas of robotics can harness Epically Powerful to get robotic systems running quickly, supported by vetted hardware and armed with a software backbone that can be easily deployed and customized for domain-specification applications.

## VI. CONCLUSION

Epically Powerful provides a modular mechatronics infrastructure and open-source software package to build and operate wearable robotic systems [24], [25]. The recommended mechatronics build enables users to choose from common and broadly utilized off-the-shelf actuators, single-board computers, power sources, and sensors, allowing customization based on the application. The software package easily integrates with these components and provides an easy-to-use Python interface for actuator communication, sensor data querying, data recording, visualization, and other necessary functionality. Broadly, Epically Powerful simplifies system construction and baseline operation to enable efficient and robust development of custom wearable robotic systems and beyond.

## APPENDIX A

The following table includes all devices that are running Epically Powerful at the time of this publication. The diversity of devices highlights the flexibility and customizability of the hardware configuration while still being able to use Epically Powerful's software architecture as the controls backbone.

| Affiliation | Device | Computer | Actuator(s) | EP-Integrated Sensor(s) | Battery | Operating Frequency | Added Sensor Peripherals | Software Additions |
|---|---|---|---|---|---|---|---|---|
| Georgia Tech EPIC/PoWeR Labs | Hip Exo (2 DOF) [12] | NVIDIA Jetson Orin Nano | AK80-9 (4) | MicroStrain IMUs (8) | 24V 2Ah drill | 200 Hz | Force sensitive resistors (4) | Real-time deep learning models (2) |
| Georgia Tech EPIC/PoWeR Labs | Knee Exo | NVIDIA Jetson Orin Nano | AK10-9 V2 (2) | MicroStrain IMUs (6) | 24V 4Ah drill | 200 Hz | XSENSOR Insoles (2) | Real-time deep learning model (1) |
| Georgia Tech EPIC Lab | Knee Exo | Raspberry Pi 4B+ | CyberGear Micromotor (1) | MicroStrain IMUs (2) | 20V 1.5Ah drill | 200 Hz | Force sensitive resistors (1) | None |
| Georgia Tech EPIC Lab | Hip Exo (1 DOF) | NVIDIA Jetson Orin Nano | AK 80-9 (2) | Microstrain IMUs (5) | 24V 2Ah drill | 200 Hz | XSENSOR Insoles (2) | Real-time deep learning model (1) |
| Georgia Tech EPIC/PoWeR Labs | Ankle Exo | NVIDIA Jetson Orin Nano | AK80-9 (2) OR AK80-9 (2) + Micromotor (2) | MicroStrain IMUs (5) | 24V 2 Ah drill | 200 Hz | XSENSOR insoles (2), load cells (2) | Real-time deep learning models (2) |
| Georgia Tech EPIC Lab | Hip-Knee Exo | NVIDIA Jetson Orin Nano | AK80-9 (4) | MicroStrain IMUs (5) | 20V 3Ah drill (2) | 200 Hz | XSENSOR Insoles (2) | Real-time deep learning model (1) |
| Georgia Tech EPIC/PoWeR Labs | Sensor Suit [35] | NVIDIA Jetson Orin Nano | None | MicroStrain IMUs (6) | 24V 2Ah drill | 200 Hz | XSENSOR Insoles (2) | Real-time deep learning model (1) |
| Georgia Tech EPIC Lab | Sensor Suit | Raspberry Pi 5 | None | MicroStrain IMUs (5) | 5V power bank | 200 Hz | XSENSOR Insoles (2) | None |
| Northeastern Shepherd Lab | Hip Exo (1 DOF) | NVIDIA Jetson Orin Nano | AK80-9 (2) OR AK10-9 (2) | None | 20V 5Ah drill | 200 Hz | None | None |
| Carnegie Mellon MetaMobility Lab | Hip Exo (1 DOF) | NVIDIA Jetson Orin Nano | AK80-9 (2) | None | HRB 24V 3.3 Ah LiPo | 200 Hz | TDK InvenSense IMU | Real-time deep learning model (2) |
| U of Washington Ingraham Lab | Hip Exo (1 DOF) | Raspberry Pi 5 | AK80-9 V2.0 (2) | None | 24V 2Ah drill | 100 Hz | Teensy + 6DOF IMU (2) | None |

## ACKNOWLEDGMENT

The authors would like to thank Dr. Max Shepherd, Fatima Tourk, and other members of the Northeastern Shepherd Lab, Dr. Inseung Kang, Nate Shoemaker-Trejo, Rajiv Joshi, and other members of the Carnegie Mellon MetaMobility Lab, and Dr. Kim Ingraham, Zijie Jin, and other members of the Ingraham Lab, as well as the members of the Georgia Tech EPIC and PoWeR Labs for their feedback and alpha stage testing of Epically Powerful hardware and software.

## REFERENCES

[1] Dephy, Inc., "Dephy ExoBoot." Accessed: Sep. 16, 2025. [Online]. Available: https://www.dephy.com/
[2] Biomotum Inc., "Biomotum SPARK." Accessed: Sep. 16, 2025. [Online]. Available: https://www.biomotum.com/
[3] Verve, Inc., "Verve Motion SafeLift." Accessed: Sep. 16, 2025. [Online]. Available: https://vervemotion.com/
[4] Skip, "Skip MO/GO." Accessed: Sep. 16, 2025. [Online]. Available: https://www.skipwithjoy.com/
[5] Roam Robotics, "Roam Robotic Knee Brace." Accessed: Sep. 16, 2025. [Online]. Available: https://www.roamrobotics.com/
[6] Hypershell, "Hypershell X Series." Accessed: Oct. 03, 2025. [Online]. Available: https://hypershell.tech/
[7] T. Zhang, M. Tran, and H. Huang, "Design and Experimental Verification of Hip Exoskeleton With Balance Capacities for Walking Assistance," *IEEEASME Trans. Mechatron.*, vol. 23, no. 1, pp. 274–285, Feb. 2018, doi: 10.1109/TMECH.2018.2790358.
[8] C. Nesler, G. Thomas, N. Divekar, E. J. Rouse, and R. D. Gregg, "Enhancing Voluntary Motion With Modular, Backdrivable, Powered Hip and Knee Orthoses," *IEEE Robot. Autom. Lett.*, vol. 7, no. 3, pp. 6155–6162, Jul. 2022, doi: 10.1109/LRA.2022.3145580.
[9] M. K. Ishmael, D. Archangeli, and T. Lenzi, "A Powered Hip Exoskeleton With High Torque Density for Walking, Running, and Stair Ascent," *IEEEASME Trans. Mechatron.*, vol. 27, no. 6, pp. 4561–4572, Dec. 2022, doi: 10.1109/TMECH.2022.3159506.
[10] P. Slade, M. J. Kochenderfer, S. L. Delp, and S. H. Collins, "Personalizing exoskeleton assistance while walking in the real world," *Nature*, vol. 610, no. 7931, Art. no. 7931, Oct. 2022, doi: 10.1038/s41586-022-05191-1.
[11] F. Sup, A. Bohara, and M. Goldfarb, "Design and Control of a Powered Transfemoral Prosthesis," *Int. J. Robot. Res.*, vol. 27, no. 2, pp. 263–273, Feb. 2008, doi: 10.1177/0278364907084588.
[12] J. K. Leestma, S. Mathur, M. D. Anderton, G. S. Sawicki, and A. J. Young, "Dynamic Duo: Design and Validation of an Autonomous Frontal and Sagittal Actuating Hip Exoskeleton for Balance Modulation During Perturbed Locomotion," *IEEE Robot. Autom. Lett.*, vol. 9, no. 5, pp. 3995–4002, May 2024, doi: 10.1109/LRA.2024.3371290.
[13] Z. F. Lerner *et al.*, "An Untethered Ankle Exoskeleton Improves Walking Economy in a Pilot Study of Individuals With Cerebral Palsy," *IEEE Trans. Neural Syst. Rehabil. Eng.*, vol. 26, no. 10, pp. 1985–1993, Oct. 2018, doi: 10.1109/TNSRE.2018.2870756.
[14] A. T. Asbeck, S. M. M. De Rossi, K. G. Holt, and C. J. Walsh, "A biologically inspired soft exosuit for walking assistance," *Int. J. Robot. Res.*, vol. 34, no. 6, pp. 744–762, May 2015, doi: 10.1177/0278364914562476.
[15] P. M. Wensing, A. Wang, S. Seok, D. Otten, J. Lang, and S. Kim, "Proprioceptive Actuator Design in the MIT Cheetah: Impact Mitigation and High-Bandwidth Physical Interaction for Dynamic Legged Robots," *IEEE Trans. Robot.*, vol. 33, no. 3, pp. 509–522, Jun. 2017, doi: 10.1109/TRO.2016.2640183.
[16] G. Bledt, M. J. Powell, B. Katz, J. Di Carlo, P. M. Wensing, and S. Kim, "MIT Cheetah 3: Design and Control of a Robust, Dynamic Quadruped Robot," in *2018 IEEE/RSJ International Conference on Intelligent Robots and Systems (IROS)*, Oct. 2018, pp. 2245–2252. doi: 10.1109/IROS.2018.8593885.
[17] A. Bajpai *et al.*, "Design and Validation of a Versatile High Torque Quasidirect Drive Hip Exoskeleton," *IEEEASME Trans. Mechatron.*, vol. 29, no. 1, pp. 789–797, Feb. 2024, doi: 10.1109/TMECH.2023.3334795.
[18] S. Chen *et al.*, "Wearable Knee Assistive Devices for Kneeling Tasks in Construction," *IEEEASME Trans. Mechatron.*, vol. 26, no. 4, pp. 1989–1996, Aug. 2021, doi: 10.1109/TMECH.2021.3081367.
[19] S. Yu *et al.*, "Quasi-Direct Drive Actuation for a Lightweight Hip Exoskeleton With High Backdrivability and High Bandwidth," *IEEEASME Trans. Mechatron.*, vol. 25, no. 4, pp. 1794–1802, Aug. 2020, doi: 10.1109/TMECH.2020.2995134.
[20] H. Zhu, C. Nesler, N. Divekar, V. Peddinti, and R. D. Gregg, "Design Principles for Compact, Backdrivable Actuation in Partial-Assist Powered Knee Orthoses," *IEEEASME Trans. Mechatron.*, vol. 26, no. 6, pp. 3104–3115, Dec. 2021, doi: 10.1109/TMECH.2021.3053226.
[21] J.-F. Duval and H. M. Herr, "FlexSEA: Flexible, Scalable Electronics Architecture for wearable robotic applications," in *2016 6th IEEE International Conference on Biomedical Robotics and Biomechatronics (BioRob)*, Jun. 2016, pp. 1236–1241. doi: 10.1109/BIOROB.2016.7523800.
[22] A. F. Azocar, L. M. Mooney, J.-F. Duval, A. M. Simon, L. J. Hargrove, and E. J. Rouse, "Design and clinical implementation of an open-source bionic leg," *Nat. Biomed. Eng.*, vol. 4, no. 10, pp. 941–953, Oct. 2020, doi: 10.1038/s41551-020-00619-3.

[23] J. R. Williams *et al.*, "OpenExo: An open-source modular exoskeleton to augment human function," *Sci. Robot.*, vol. 10, no. 103, p. eadt1591, Jun. 2025, doi: 10.1126/scirobotics.adt1591.

[24] "gatech-epic-power/epically-powerful: A python toolbox for controlling and handling actuators and IMUs, commonly used in robotics." Accessed: Nov. 07, 2025. [Online]. Available: https://github.com/gatech-epic-power/epically-powerful

[25] "Epically Powerful Documentation." Accessed: Nov. 07, 2025. [Online]. Available: https://gatech-epic-power.github.io/epically-powerful/

[26] F. M. Tourk, B. Galoaa, S. Shajan, A. J. Young, M. Everett, and M. K. Shepherd, "Uncertainty-Aware Ankle Exoskeleton Control," Aug. 28, 2025, *arXiv*: arXiv:2508.21221. doi: 10.48550/arXiv.2508.21221.

[27] D. D. Molinaro, K. L. Scherpereel, E. B. Schonhaut, G. Evangelopoulos, M. K. Shepherd, and A. J. Young, "Task-agnostic exoskeleton control via biological joint moment estimation," *Nature*, vol. 635, no. 8038, pp. 337–344, Nov. 2024, doi: 10.1038/s41586-024-08157-7.

[28] "SocketCAN - Controller Area Network — The Linux Kernel documentation." Accessed: Oct. 03, 2025. [Online]. Available: https://docs.kernel.org/networking/can.html

[29] "python-can 4.6.1 documentation." Accessed: Nov. 07, 2025. [Online]. Available: https://python-can.readthedocs.io/en/stable/index.html

[30] *LORD-MicroStrain/MSCL*. (Nov. 03, 2025). C++. MicroStrain by HBK. Accessed: Nov. 07, 2025. [Online]. Available: https://github.com/LORD-MicroStrain/MSCL

[31] M. K. Shepherd, D. D. Molinaro, G. S. Sawicki, and A. J. Young, "Deep Learning Enables Exoboot Control to Augment Variable-Speed Walking," *IEEE Robot. Autom. Lett.*, vol. 7, no. 2, pp. 3571–3577, Apr. 2022, doi: 10.1109/LRA.2022.3147565.

[32] S. Cheng, C. A. Laubscher, T. K. Best, and R. D. Gregg, "Ambilateral Activity Recognition and Continuous Adaptation With a Powered Knee-Ankle Prosthesis," *IEEE Trans. Robot.*, vol. 41, pp. 2251–2267, 2025, doi: 10.1109/TRO.2025.3539206.

[33] M. Quigley *et al.*, "ROS: an open-source Robot Operating System," *ICRA Workshop Open Source Softw.*, 2009.

[34] "PlotJuggler," PlotJuggler. Accessed: Nov. 07, 2025. [Online]. Available: https://plotjuggler.io

[35] R. T. F. Casey *et al.*, "The Second Skin: A Wearable Sensor Suite that Enables Real-Time Human Biomechanics Tracking Through Deep Learning," *IEEE Trans. Biomed. Eng.*, pp. 1–10, 2025, doi: 10.1109/TBME.2025.3589996.

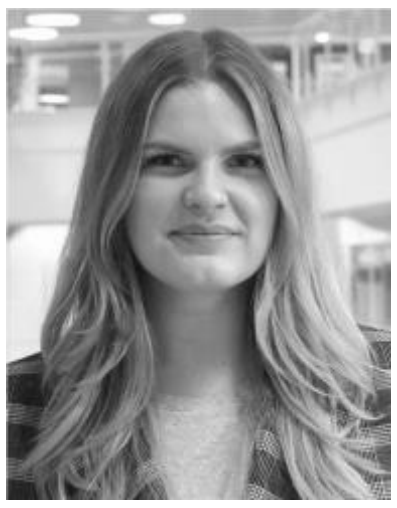

**JENNIFER K. LEESTMA** received the B.S. degree in biomedical engineering from the University of Wisconsin-Madison, Madison, WI, USA in 2018, the M.S. degree in mechanical engineering from the Georgia Institute of Technology, Atlanta, GA, USA in 2022, and the Ph.D. degree in robotics from the Georgia Institute of Technology, Atlanta, GA, USA in 2024. Since 2024, she has been a postdoctoral fellow in bioengineering in the School of Engineering and Applied Sciences at Harvard University, Cambridge, MA, USA. Her research interests include restoring and augmenting sensory and motor function using neurally-integrated robotic systems.

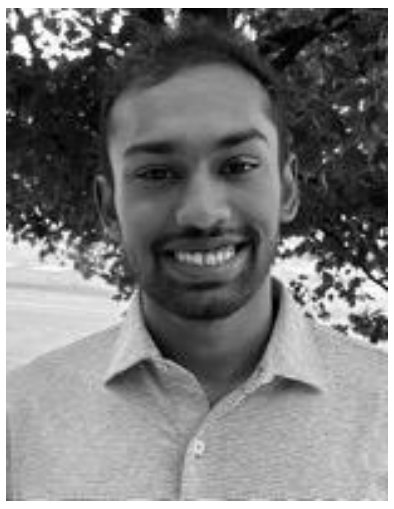

**SIDDHARTH R. NATHELLA** received the B.S. degree in mechanical engineering from Purdue University, West Lafeyette, IN, USA, in 2022. He is currently pursuing the Ph.D. degree in mechanical engineering at the Georgia Institute of Technology, Atlanta, GA, USA. His research interests include gait training and rehabilitation using biofeedback and wearable robotics.

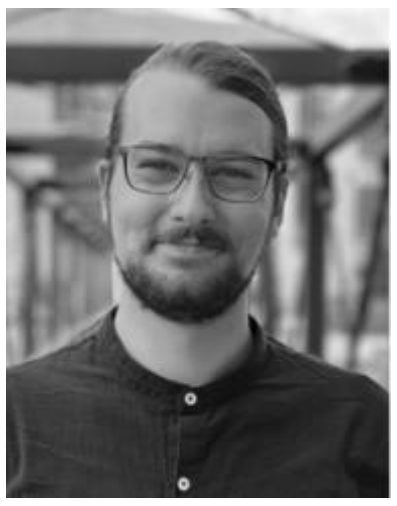

**CHRISTOPH P. O. NUESSLEIN** received the B.S. degree in mechanical engineering from the University of Massachusetts-Amherst, Amherst, MA, USA, in 2021. He is currently pursuing the Ph.D. degree in robotics at the Georgia Institute of Technology, Atlanta, GA, USA. His research interests include reducing joint injury prevalence through exoskeleton assistance in manual labor environments.

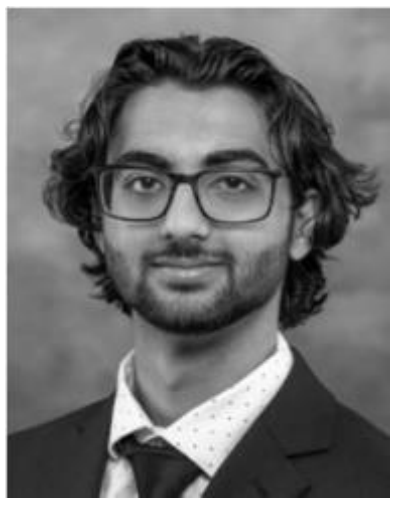

**SNEHIL MATHUR** (S’24) received the B.S. degree in electrical engineering from the Georgia Institute of Technology, Atlanta, GA, USA in 2024. He is currently pursuing the Ph.D. degree in mechanical engineering from Rice University, Houston, TX, USA. His research interests include functional and teleoperated upper-limb rehabilitation through wearable robots and multimodal sensing.

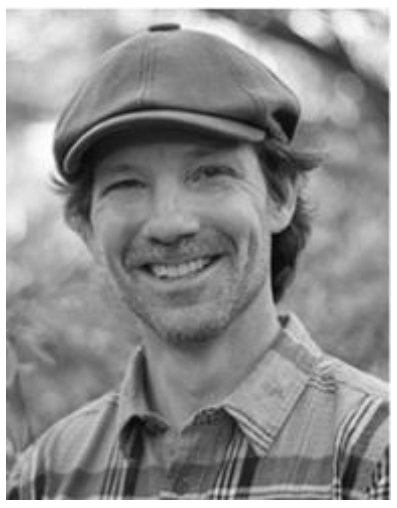

**GREGORY S. SAWICKI** (M’21) received the B.S. degree in mechanical engineering from Cornell University, Ithaca, NY, USA in 1999**,** the M.S. degree in mechanical engineering from University of California-Davis, Davis, CA, USA in 2001, and the Ph.D. degree in human neuromechanics from the University of Michigan, Ann Arbor, MI, USA in 2007. He did his postdoctoral fellowship in integrative biology at Brown University, Providence, RI, USA. From 2009 to 2017, he was an Associate Professor in the Joint Department of Biomedical Engineering at the University of North Carolina at Chapel Hill, Chapel Hill, NC, USA, and North Carolina State University, Raleigh, NC, USA. He is currently a Professor in the George W. Woodruff School of Mechanical Engineering and the School of Biological Sciences at the Georgia Institute of Technology, Atlanta, GA, USA. His research focuses on discovering physiological principles underpinning locomotion performance and applying them to develop lower-limb robotic devices capable of improving both healthy and impaired human locomotion.

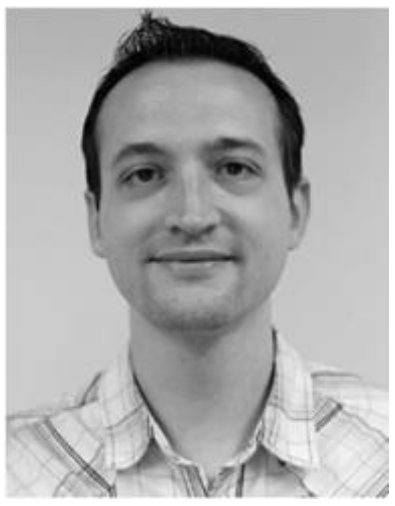

**AARON J. YOUNG** (S’12–M’16–SM’23) received the B.S. degree in biomedical engineering from Purdue University, West Lafeyette, IN, USA in 2009, the M.S. degree in biomedical engineering from Northwestern University, Evanston, IL, USA in 2011, and the Ph.D. degree in biomedical engineering from Northwestern University, Evanston, IL, USA in 2014. He did his postdoctoral fellowship in mechanical engineering at the University of Michigan, Ann Arbor, MI, USA. He is currently an Associate Professor with the George W. Woodruff School of Mechanical Engineering at the Georgia Institute of Technology, Atlanta, GA, USA. His research focuses on developing and deploying machine learning-driven control approaches for wearable robotic systems, aimed at improving and enhancing human mobility.